\documentclass{article}
\usepackage{arxiv_preprint,times}
\usepackage[T1]{fontenc}
\usepackage[utf8]{inputenc}
\usepackage{microtype}
\usepackage{graphicx}
\usepackage{booktabs,array,tabularx,longtable,multirow}
\usepackage{caption}
\usepackage{amsmath,amssymb,mathtools}
\usepackage[hidelinks]{hyperref}
\usepackage{xurl}
\usepackage[capitalize,noabbrev]{cleveref}
\graphicspath{{../figures/}}
\newcommand{\method}{\textsc{LARDM}}
\newcommand{\attnres}{\textsc{AttnRes}}
\newcommand{\Read}{\operatorname{Read}}
\newcommand{\RMS}{\operatorname{RMSNorm}}

\newcommand{\stageL}{\mathrm{L}}
\newcommand{\stageR}{\mathrm{R}}

\title{\Large What Survives a Loop?\\A Shared Decision Model for Encoders and Decoders}
\author{%
Yixin Peng$^{1\star}$, and Stefan Decker$^{1,2}$ \\
$^{1}$ RWTH Aachen University, Aachen, Germany \\
\texttt{\{peng,decker\}@dbis.rwth-aachen.de} \\
$^{2}$ Fraunhofer FIT, Sankt Augustin, Germany}
\hypersetup{
  pdftitle={What Survives a Loop? A Shared Decision Model for Encoders and Decoders},
  pdfauthor={Yixin Peng, Diego Collarana, Stefan Decker}
}

\begin{document}
\raggedbottom
\maketitle
\begingroup
\renewcommand{\thefootnote}{\ensuremath{\star}}
\footnotetext[0]{Corresponding author: Yixin Peng
(\href{mailto:peng@dbis.rwth-aachen.de}{\texttt{peng@dbis.rwth-aachen.de}}).}
\endgroup

\begin{abstract}
A looped Transformer refines token representations by applying the same layers
across successive passes. With additive residuals, each pass receives a single
accumulated state per token. Attending to every earlier sublayer output would let the
model reweight individual updates, but each pass would lengthen the list it
must read. We investigate whether bidirectional encoders and causal decoders
benefit equally from access to this history when selecting a commonsense
answer. Our Looped Attention Residuals model lets each sublayer weight earlier
outputs from the current pass individually and read a separate tokenwise sum
from each completed pass. We pretrain eight
variants on the same 3B FineWeb-Edu tokens, varying the backbone, the number of
passes (one or four), and the residual connection. All models score the five
CommonsenseQA candidates with the same decision rule. Before supervised
fine-tuning, we compare learning from concept descriptions, ranking candidate
answers to relation questions, and applying the two stages in sequence.
With four passes and direct fine-tuning, replacing additive residuals with
depth reads raises held-out accuracy by $10.23$ percentage points for the
encoder and $0.96$ for the decoder. Relation ranking alone improves on direct
fine-tuning for all eight variants and six released baselines. Modeling
descriptions before ranking brings the Looped Attention Residuals encoder to
$50.36\%$. Both looped decoders outperform GPT-Neo-125M on all four training
routes with roughly one hundredth of its reported pretraining token exposure.
\end{abstract}

\section{Introduction}
\label{sec:introduction}

A waterfall and a bridge are both associated with a river. In one
CommonsenseQA example, catching falling water in a cup points to the
waterfall. In another, watching the water fall while staying dry points to
the bridge \citep{talmor2019commonsenseqa}. The association with a river is
useful in both cases. Which answer fits depends on how the candidate relates
to the rest of the question. A looped Transformer can process each
question--candidate pair several times with the same layers. When those
layers run again, which parts of their earlier computation should they be
able to revisit?

Layer reuse and residual connections govern different parts of this
decision. Reusing a stack gives the model more steps to refine a token's
representation \citep{dehghani2019universal,zhu2025looped}. With additive
residuals, each pass hands the next one a single accumulated vector per
token. Attention Residuals lets a sublayer weight individual earlier outputs
\citep{kimi2026attention}, but retaining all of them would lengthen the list
on every pass. We use the end of a pass as the point at which to summarize.
Within the current pass, a sublayer can weight each preceding output
individually. Once that pass ends, we sum its outputs at each token and
keep the sum as a separate source. On the fourth pass, for example, a
sublayer can give different weights to the first three passes and still
choose among individual updates from the fourth (\cref{fig:looped}).

The same connection gives an encoder and a decoder access to different
information. When the candidate follows the question, a bidirectional
encoder can represent a question word in light of that candidate. A causal
decoder encounters the word before the candidate appears. The final token
can see the entire pair in both cases. We therefore score that token with
the same kind of linear head in each backbone and train both to choose among
five answers. We cross one or four passes with additive or attention-based
residuals in each backbone family. All eight models share a tokenizer,
stack dimensions, and a one-epoch budget of 3B FineWeb-Edu content tokens
\citep{penedo2024fineweb}. The four-pass models spend more computation on
each example.

Answer selection also depends on how the model encounters commonsense
knowledge before fine-tuning. Predicting the words in a definition gives
the model evidence about a concept. Choosing which of five concepts
completes a relation gives it practice using such evidence to make a
decision. We compare direct fine-tuning with each form of preparation and
with the sequence from description modeling to relation ranking. Applying
the same routes to six released checkpoints puts the controlled
architecture comparisons in the context of other pretraining corpora and
budgets. Three findings emerge on the CommonsenseQA in-house test set
(\cref{tab:main}).

\begin{itemize}
    \item \textbf{The encoder gains more from depth reads.}
    With four passes and direct fine-tuning, the encoder rises from
    $35.46\%$ with additive residuals to $45.69\%$ with depth reads.
    The decoder starts at $45.45\%$ and reaches $46.41\%$. The two models
    finish within one point of each other despite the difference in gains.
    \item \textbf{Relation ranking transfers across architectures.}
    Relation ranking alone improves held-out accuracy over direct
    fine-tuning for all fourteen models. Adding description modeling
    before ranking brings the looped depth-read encoder to $50.36\%$,
    its highest accuracy. The corresponding decoder scores higher with
    relation ranking alone.
    \item \textbf{A 3B-token run surpasses selected released checkpoints.}
    Both looped decoders exceed GPT-Neo-125M on all four matched routes,
    although GPT-Neo reports 300B pretraining tokens
    \citep{eleuther2021gptneo}. OPT-125M remains ahead of our decoders,
    and the released encoders lead our encoders on every matched route.
\end{itemize}

\section{Background and Related Work}
\label{sec:related}

\subsection{Commonsense Answer Selection in Encoders and Decoders}
\label{sec:problem}

CommonsenseQA uses ConceptNet relations to elicit questions whose answer
choices include several concepts related to the same source
\citep{speer2017conceptnet,talmor2019commonsenseqa}. Let
$\mathcal D=\{(q_i,\mathcal C_i,y_i)\}_{i=1}^{N}$ denote the resulting
decision task, where $\mathcal C_i=(c_{i1},\ldots,c_{i5})$ contains the
candidate answers and $y_i$ indexes the correct one. Each candidate receives
a score $s_\theta(q_i,c_{ij})$, and the model predicts
$\hat y_i=\arg\max_j s_\theta(q_i,c_{ij})$. This formulation leaves open how
to score a pair.

One option is to use the language model itself. A causal model assigns a
continuation a probability from left to right. A masked language model can
score a sequence by masking each token in turn and aggregating its
conditional log probabilities \citep{salazar2020masked}. These scores
condition on different context. A supervised scalar head gives both families
the same decision rule. In our comparison, each backbone learns to score the
last token of a complete pair, and all five scores enter the same
classification loss (\cref{sec:decision}).

The pretrained representations supplied to that head still differ.
BERT and RoBERTa learn from masked tokens with bidirectional context, and
DeBERTa separates content and position in its attention mechanism
\citep{devlin2019bert,liu2019roberta,he2021deberta}. GPT-2, GPT-Neo, and
OPT use causal next-token prediction
\citep{radford2019language,eleuther2021gptneo,zhang2022opt}. Their released
checkpoints also differ in training text and exposure. We use them as
downstream reference points. For the architectural comparisons we train
encoders and decoders on the same input tokens, keeping each family's
attention mask and language-model objective together.

\subsection{Repeated Computation and Access to Earlier Layers}

The Universal Transformer refines token states by applying a shared
transformation repeatedly and can learn when individual positions should
stop updating \citep{dehghani2019universal}. Ouro similarly studies how
a looped language model can learn to allocate computation across depth
\citep{zhu2025looped}. The number of passes determines how many updates
a model can make. Even with that number fixed, there is a choice about
which earlier states can contribute directly to the next update.

Earlier work on translation gave the decoder access to a learned mixture
of encoder layers \citep{bapna2018transparent}. Attention Residuals makes
layer selection part of the stack itself. At each token position, Full
Attention Residuals weights preceding sublayer outputs, while the Block
variant combines outputs into groups to reduce the number of sources
\citep{kimi2026attention}. Our design ties the grouping to weight reuse.
For example, the output of layer three is individually accessible to layer
four on the same pass. On the next pass its contribution is part of the
previous pass's sum. This preserves a separate source for each completed
pass and keeps at most one pass's worth of individual sublayer outputs available to any read. The additive loop with the same
number of passes provides the control for this choice.

\subsection{Knowledge in the Input and in Post-Training}

Knowledge graphs can supply evidence at prediction time. GreaseLM exchanges
information between language and graph representations while processing
a question and candidate \citep{zhang2022greaselm}. A graph can also supply
training targets. For example, \citet{kim2022modularized} convert
commonsense graphs into synthetic question-answering examples and study
zero-shot transfer across knowledge sources. Our relation-ranking stage
likewise turns graph facts into choices. Its role here is to initialize
a scorer that will subsequently be fine-tuned on labeled CommonsenseQA
questions. At prediction time, it needs only the question and candidate text.

Continued language modeling offers another way to use task-relevant
knowledge. \citet{gururangan2020adapt} show that further pretraining on
domain or task text can improve downstream classification. We apply this
principle to definitions and example sentences for concepts in the task.
The two kinds of intermediate training transfer different parameters. Description modeling initializes the backbone, while relation ranking also
trains the scalar head used for answer selection. Testing each stage alone
and then descriptions followed by ranking lets us ask whether learning
about a concept in text adds to the benefit of practicing a five-way choice.

\section{Method}
\label{sec:method}

Our Looped Attention Residuals Decision Model (\method{}, labeled Looped
\attnres{} in the experiments) applies a shared stack of layers to each
question--candidate pair. Both backbone variants implement the following
scoring rule.

\subsection{A Shared Decision Interface}
\label{sec:decision}

For each candidate $c_j$ and question $q$, both backbones receive
\begin{equation}
 x_j=[\mathrm{BOS};\ \texttt{Question: }q;\ \texttt{Answer: }c_j;\ \mathrm{EOS}],
 \label{eq:input}
\end{equation}
where the question and answer fields are separated by a newline.
Let $E\in\mathbb R^{n\times d}$ be its token embeddings after embedding dropout,
with sequence length $n$ and width $d$. We use $D$ independently parameterized
Transformer layers, each containing an attention and a feed-forward sublayer,
and execute the stack $T$ times. Layer parameters differ across the $D$ layers
but are shared across traversals. Our principal configuration is $D=12$, $T=4$.

Let $H_j$ denote the final token representations and $e_j$ the position of
the last valid EOS token. Each model independently encodes the candidates
and scores them with a learned vector $v\in\mathbb R^d$ according to
\begin{equation}
 s_j=v^\top H_{j,e_j},\qquad
 p_j=\frac{\exp(s_j)}{\sum_{k=1}^{K}\exp(s_k)},\qquad
 \mathcal L_{\mathrm{rank}}=-\log p_y.
 \label{eq:decision}
\end{equation}
The softmax couples the five scores within a question. We fine-tune the
backbone and its scorer jointly. Both backbone families use this same
decision rule, with independently learned scorer weights.

\subsection{From Residual Accumulation to Selective Depth Reads}
\label{sec:depth}

\paragraph{Standard residuals.}
Writing $h$ for a layer's input, a standard PreNorm layer computes
\begin{align}
 a_\ell &= \operatorname{Attention}_\ell(\RMS_{\ell,a}(h)), &
 \widetilde h &= h+a_\ell, \nonumber\\
 m_\ell &= \operatorname{FFN}_\ell(\RMS_{\ell,m}(\widetilde h)), &
 h' &= \widetilde h+m_\ell.
 \label{eq:standard}
\end{align}
Each sublayer receives the accumulated state. A later sublayer can use an
earlier update only through the states that have incorporated it.

\begin{figure}[!t]
 \centering
 \includegraphics[width=\linewidth]{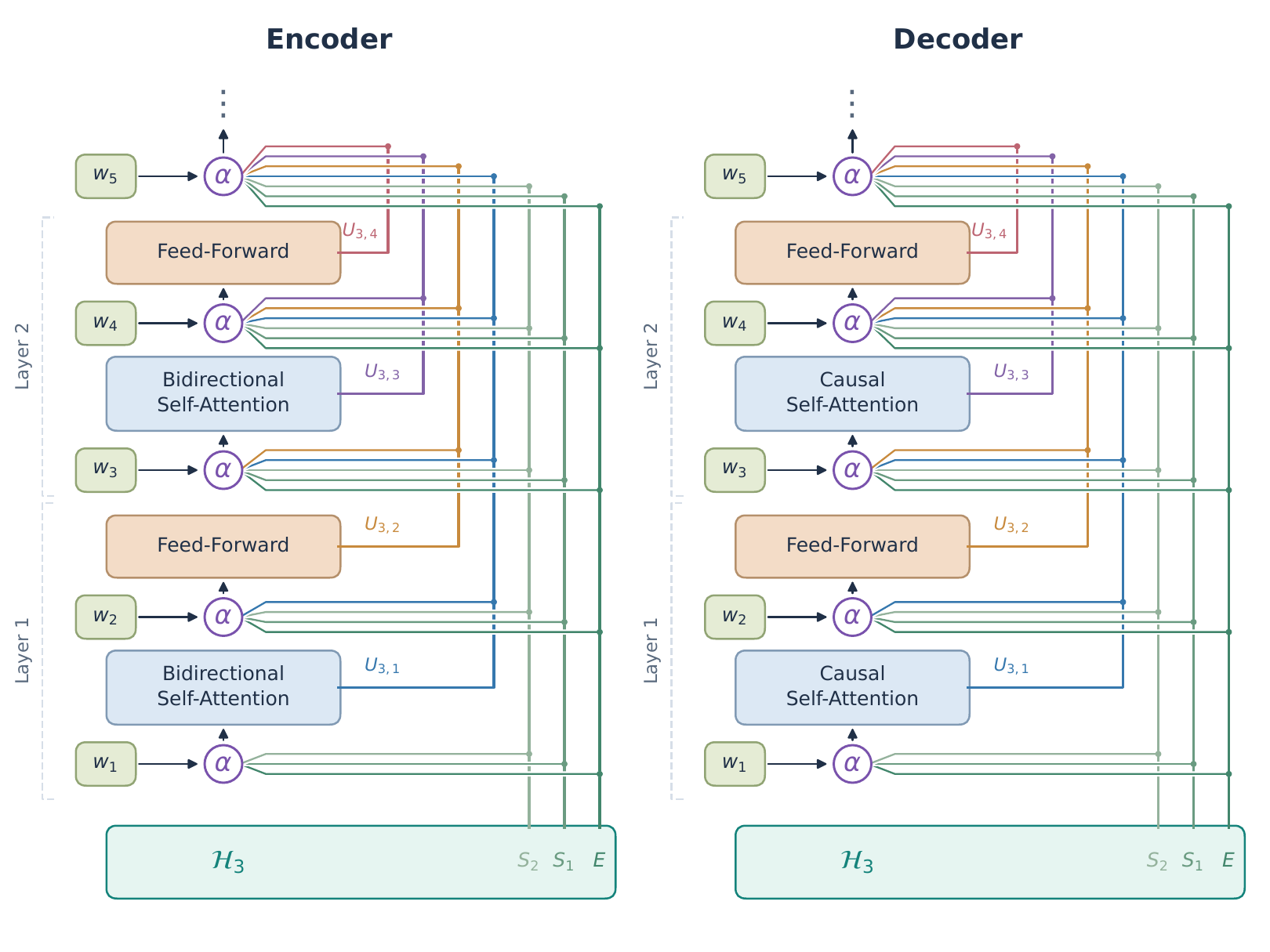}
 \caption{Depth reads inside the third traversal. Each $\alpha$ node weights
 the embedding $E$, summaries $S_1,S_2$, and the raw outputs already produced
 in this traversal. Each track carries one source. The first four reads feed
 the attention and feed-forward sublayers of layers 1 and 2, and $w_5$ feeds
 layer 3. The encoder and decoder share this wiring and differ in the token
 visibility of their self-attention sublayers.}
 \label{fig:within-loop}
\end{figure}

\paragraph{Reading earlier sublayer outputs.}
Attention Residuals \citep{kimi2026attention} makes previous sublayer outputs
individually accessible. Before each attention or feed-forward transformation,
the model computes a weighted sum of the sources available at that depth.
\Cref{fig:within-loop} illustrates this operation during a loop.

\paragraph{Attention over depth sources.}
For a source list $\mathcal S=(V_1,\ldots,V_m)$ with
$V_i\in\mathbb R^{n\times d}$, define the read at sublayer position $j$ by
\begin{align}
 b_{j,i,t} &= w_j^\top\RMS^{\mathrm{key}}_j(V_{i,t}), &
 \alpha_{j,i,t} &= \frac{\exp(b_{j,i,t})}{\sum_{k=1}^{m}\exp(b_{j,k,t})},
 \label{eq:weights}\\
 \Read_j(\mathcal S)_t &= \sum_{i=1}^{m}\alpha_{j,i,t}V_{i,t}.
 \label{eq:read}
\end{align}
Each read position $j$ has its own query $w_j\in\mathbb R^d$ and learned
key-normalization scale. The values are the raw source vectors. Initializing
the query to zero assigns equal weight to the available sources at the start
of training. As the query learns, the weight of a source depends on its
content at token position $t$.

The two attention operations mix information along different axes.
Self-attention combines token positions under the backbone's visibility
mask. A depth read combines earlier representations of a single position. A
causal decoder can therefore use the same depth-read rule as a bidirectional
encoder while preserving its left-to-right dependency structure.

\paragraph{Full Attention Residuals.}
In a single traversal, sublayer $j\in\{1,\ldots,2D\}$ reads
$\mathcal S_j=(E,U_1,\ldots,U_{j-1})$ and produces
\begin{equation}
 U_j=F_j\!\left(\RMS^{\mathrm{pre}}_j(\Read_j(\mathcal S_j))\right),
 \label{eq:sublayer}
\end{equation}
where $F_j$ alternates between token attention and the feed-forward network.
The model appends $U_j$ to the history. A separate output query reads
$(E,U_1,\ldots,U_{2D})$ before final RMS normalization. We use this
single-pass model as the Full \attnres{} baseline.

\subsection{What Crosses a Loop Boundary?}
\label{sec:looped}

\paragraph{Detailed reads within each traversal.}
For traversal $r\in\{1,\ldots,T\}$, let $U_{r,j}$ denote its $j$th raw sublayer
output. Before computing this output, the source list is
\begin{equation}
 \mathcal S_{r,j}=
 (E,S_1,\ldots,S_{r-1},U_{r,1},\ldots,U_{r,j-1}),
 \label{eq:history}
\end{equation}
and
\begin{equation}
 U_{r,j}=F_j\!\left(\RMS^{\mathrm{pre}}_j
       (\Read_j(\mathcal S_{r,j}))\right).
 \label{eq:looped}
\end{equation}
All parameters of $F_j$, both normalization operations, and $w_j$ are shared
across $r$. The same transformations therefore operate on a changing history.

\paragraph{Summaries across traversals.}
After a non-final traversal, we form
\begin{equation}
 S_r=\sum_{j=1}^{2D}U_{r,j}\in\mathbb R^{n\times d}.
 \label{eq:summary}
\end{equation}
The sum includes only the outputs produced on pass $r$. At the start of the
next pass, $E,S_1,\ldots,S_r$ are available as separate sources. A later
sublayer can thus reweight entire completed passes and individual outputs
from its own pass. Gradients flow through each summary to its constituent
sublayers.

\begin{figure}[t]
 \centering
 \includegraphics[width=\linewidth]{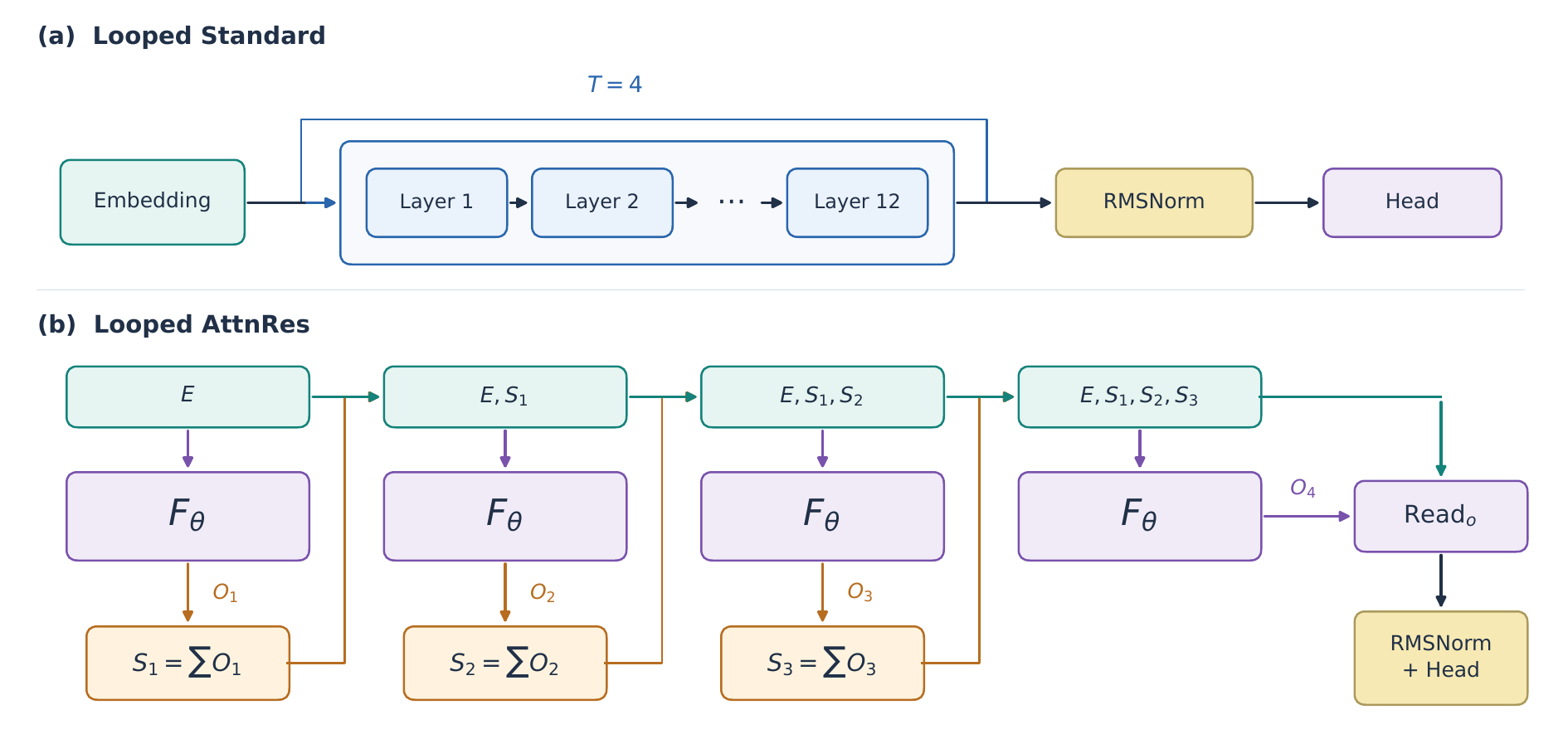}
 \caption{What survives each traversal of a 12-layer stack. (a) Looped
 Standard passes one accumulated state through the stack four times.
 (b) Looped \attnres{} retains one tokenwise sum $S_r$ of the 24 raw sublayer
 outputs $O_r$ from each completed traversal. On the fourth pass the final
 query reads $E,S_1,S_2,S_3$ and all 24 outputs in $O_4$. The normalized
 representation feeds the language-model head or the EOS answer scorer.}
 \label{fig:looped}
\end{figure}

\paragraph{Final readout.}
The output query reads the full local history of the final traversal,
\begin{equation}
 H=\RMS^{\mathrm{out}}\!\left(
   \Read_o(E,S_1,\ldots,S_{T-1},U_{T,1},\ldots,U_{T,2D})\right).
 \label{eq:final}
\end{equation}
The output read has its own query and key-normalization parameters.
With $T=1$, this construction reduces to Full \attnres{} under the same weights.
\Cref{fig:within-loop} expands the shared stack, while \cref{fig:looped}
shows how its raw outputs are summarized and reused across traversals.

\paragraph{Matched looped control.}
Looped Standard applies \cref{eq:standard} repeatedly, carrying $h$ directly
across boundaries. It embeds the sequence once and applies final normalization
after all traversals. Comparing it with Looped \attnres{} holds layer reuse
and execution depth fixed while changing how each sublayer accesses history.

\paragraph{Parameter and compute cost.}
Each of the $2D$ sublayers and the final read learns a query and a
normalization scale of width $d$. Depth reads therefore add $2(2D+1)d$
parameters for any traversal count $T$. The shared stack still executes
$DT$ layers per candidate. For length $n$ and feed-forward width proportional
to $d$, its attention and feed-forward transformations cost
$\Theta(DT(nd^2+n^2d))$ per forward pass.

The additional work of a depth read depends on how many sources it combines.
Sublayer $j$ on traversal $r$ reads $r+j-1$ sources at each token.
Counting the final read in \cref{eq:final}, the number of source vectors
read per token over the forward pass is
\begin{equation}
 C(D,T)=\sum_{r=1}^{T}\sum_{j=1}^{2D}(r+j-1)+(2D+T)
       =(DT+1)(2D+T).
 \label{eq:read-cost}
\end{equation}
Scoring and combining these vectors costs $\Theta(nd\,C(D,T))$ per
candidate. Keeping every sublayer output individually across traversals
would instead give $(DT+1)(2DT+1)$ source reads per token. With pass summaries,
both the cumulative read count and the maximum number of sources per read
fall to a fraction $(2D+T)/(2DT+1)$ of their full-history values, about
$29\%$ for our four-pass model. The largest list
occupies $\Theta(nd(2D+T))$ elements. Backpropagation also retains
intermediate tensors for the depth reads.

\subsection{Knowledge Adaptation Before Answer Selection}
\label{sec:training}

We study two ways to introduce knowledge before supervised fine-tuning
(SFT) on CommonsenseQA. Both draw on entities relevant to the questions
and answer options. They differ in what the model must predict.

\paragraph{Learning from descriptions ($\stageL$).}
A definition identifies a concept and an example illustrates its use.
We continue masked language modeling (MLM) for the encoder and causal
language modeling (CLM) for the decoder on this text. The resulting
backbone initializes the decision model with a fresh answer scorer.
\Cref{app:data} describes the two versions of the description corpus.

\paragraph{Learning from relations ($\stageR$).}
Given a fact $(a,r,b)$, we turn $(a,r)$ into a question and rank $b$ among
five possible tails. For example, $(\textit{bird},\texttt{CapableOf},\textit{fly})$
yields a question of the form \emph{What can a bird do?} with \emph{fly} as its
target. We draw four alternatives from the same relation, excluding known
positive tails (\cref{app:objectives}). The model learns to distinguish
candidate answers with the downstream objective in \cref{eq:decision}, so
both the backbone and its scorer transfer to CommonsenseQA.

\paragraph{Four training routes.}
The four routes connect base pretraining (PT) to supervised fine-tuning as follows.
\begin{equation}
\begin{aligned}
 \text{Direct}:&\quad \mathrm{PT}\rightarrow\mathrm{SFT},\\
 \stageL:&\quad \mathrm{PT}\rightarrow\stageL\rightarrow\mathrm{SFT},\\
 \stageR:&\quad \mathrm{PT}\rightarrow\stageR\rightarrow\mathrm{SFT},\\
 \stageL\rightarrow\stageR:&\quad
 \mathrm{PT}\rightarrow\stageL\rightarrow\stageR\rightarrow\mathrm{SFT}.
\end{aligned}
\label{eq:routes}
\end{equation}
For each knowledge stage, we train separate one-, two-, and three-epoch
configurations. The combined route has nine possible budget pairs, with
description modeling preceding relation ranking in each case.

\section{Experiments and Results}
\label{sec:experiments}

\subsection{Experimental Setup}
\label{sec:setup}

\paragraph{Architecture comparison.}
We train eight models by crossing backbone family (encoder or decoder),
number of traversals (one or four), and residual design (additive or
attention-based). We call the additive and depth-read single-pass models
Standard and Full \attnres{}, respectively. Their four-pass counterparts
are Looped Standard and Looped \attnres{}. All use a 12-layer stack, the
RoBERTa tokenizer, and approximately 127M parameters. Base pretraining
covers 3B FineWeb-Edu content tokens in one epoch \citep{penedo2024fineweb}.
The token budget matches exposure across models. The looped variants
execute four times as many layers per example, and MLM supervises a
smaller fraction of input positions than CLM.

\paragraph{Task and model selection.}
We use the CommonsenseQA in-house split distributed with GreaseLM
\citep{zhang2022greaselm} and fine-tune each configuration for five epochs.
We select language-model checkpoints by minimum validation loss and
ranking checkpoints by maximum validation accuracy, breaking ties by
ranking loss. For each model and adaptation route, we select the epoch
budget on the CommonsenseQA development set (IHDev, abbreviated Dev) and report the accuracy
of that configuration on the held-out in-house test set (IHTest).
\Cref{app:selection} gives the budget tie-breaks. The Direct route has
no intermediate knowledge stage.

\paragraph{Reference models.}
We apply the same adaptation routes and decision task to BERT-base, RoBERTa-base, DeBERTa-base v1, GPT-2, GPT-Neo-125M,
and OPT-125M. These results put the controlled comparisons in the context
of models with different pretraining corpora and budgets
(\cref{app:baselines}).

\subsection{Main Results}
\label{sec:results}

\begin{table}[t]
\centering
\caption{CommonsenseQA accuracy (\%) on IHDev and IHTest for the same
IHDev-selected configuration. Direct uses supervised fine-tuning alone.
The other routes first model descriptions ($\stageL$), rank relation tails
($\stageR$), or do both in sequence. Within each backbone family, released
checkpoints appear above the single rule and our models below it.
Bold marks the best score per split and route within each group.
Corpus size counts billions of tokens before repeated sampling.
$^{\dagger}$ denotes a byte-based estimate (\cref{app:baselines}).}
\label{tab:main}
\footnotesize
\setlength{\tabcolsep}{1.5pt}
\begin{tabular*}{\linewidth}{@{\extracolsep{\fill}}llrrrrrrrrr@{}}
\toprule
\multirow{2}{*}{Family} & \multirow{2}{*}{Model} & Corpus &
\multicolumn{2}{c}{Direct} & \multicolumn{2}{c}{$\stageL$} &
\multicolumn{2}{c}{$\stageR$} & \multicolumn{2}{c@{}}{$\stageL\!\rightarrow\!\stageR$} \\
\cmidrule(lr){4-5}\cmidrule(lr){6-7}\cmidrule(lr){8-9}\cmidrule(l){10-11}
& & (B tokens) & IHDev & IHTest & IHDev & IHTest & IHDev & IHTest & IHDev & IHTest \\
\midrule
\multirow{7}{*}{Encoder} & BERT-base & 3.8$^{\dagger}$ & 52.09 & 50.12 & 52.33 & 50.68 & 57.90 & 52.78 & 58.15 & 53.18 \\
 & RoBERTa-base & 38$^{\dagger}$ & 65.52 & 60.68 & 65.19 & 60.84 & 67.57 & 63.26 & 69.37 & \textbf{63.58} \\
 & DeBERTa-base v1 & 19$^{\dagger}$ & \textbf{66.09} & \textbf{61.89} & \textbf{66.91} & \textbf{62.29} & \textbf{69.37} & \textbf{63.90} & \textbf{69.70} & 62.93 \\
\cmidrule(l){2-11}
 & Standard & 3 & 33.82 & 32.15 & 42.26 & 37.07 & 41.11 & 38.36 & 45.54 & 38.84 \\
 & Full \attnres{} & 3 & 45.86 & 41.02 & 45.62 & 42.55 & 48.40 & 44.96 & 49.88 & 44.96 \\
 & Looped Standard & 3 & 37.59 & 35.46 & 48.16 & 43.92 & 45.45 & 39.56 & 48.08 & 44.00 \\
 & Looped \attnres{} & 3 & \textbf{51.19} & \textbf{45.69} & \textbf{51.60} & \textbf{46.66} & \textbf{52.25} & \textbf{49.23} & \textbf{53.15} & \textbf{50.36} \\
\noalign{\vskip 2pt}
\hline\hline
\noalign{\vskip 2pt}
\multirow{7}{*}{Decoder} & GPT-2 & 9.6$^{\dagger}$ & 30.38 & 31.83 & 43.82 & 43.35 & 49.47 & 49.80 & 49.96 & 50.52 \\
 & GPT-Neo-125M & 260$^{\dagger}$ & 44.96 & 43.51 & 45.29 & 43.35 & 48.32 & 43.76 & 47.83 & 45.12 \\
 & OPT-125M & $\sim$180 & \textbf{53.97} & \textbf{49.32} & \textbf{54.46} & \textbf{49.40} & \textbf{57.00} & \textbf{52.38} & \textbf{57.33} & \textbf{51.89} \\
\cmidrule(l){2-11}
 & Standard & 3 & 45.54 & 41.50 & 45.45 & 40.69 & 47.58 & 43.84 & 47.91 & 41.02 \\
 & Full \attnres{} & 3 & 45.29 & 41.42 & 45.29 & 40.45 & 46.60 & 44.48 & 47.75 & 43.35 \\
 & Looped Standard & 3 & 48.08 & 45.45 & 48.65 & \textbf{45.61} & 49.63 & \textbf{47.46} & 49.47 & \textbf{47.70} \\
 & Looped \attnres{} & 3 & \textbf{48.73} & \textbf{46.41} & \textbf{49.22} & 45.21 & \textbf{51.11} & 46.58 & \textbf{52.01} & 46.25 \\
\bottomrule
\end{tabular*}
\end{table}

\paragraph{Gains from layer reuse and depth reads.}
The IHTest scores under Direct in \cref{tab:main} isolate the architectural comparisons
from knowledge adaptation. Reusing the additive stack for four passes
raises accuracy by $3.31$ percentage points for the encoder and $3.95$
for the decoder. At four passes, replacing additive residuals with learned
depth reads raises the encoder from $35.46\%$ to $45.69\%$ ($+10.23$ points)
and the decoder from $45.45\%$ to $46.41\%$ ($+0.96$).
The same difference between backbones is visible after one pass, where
depth reads yield $+8.87$ points for the encoder and $-0.08$ for the
decoder. Looped \attnres{} leads our encoders on all four routes.
Among our decoders it leads only on Direct, with Looped Standard scoring
higher on each knowledge route (\cref{tab:deltas}).

\paragraph{Effect of post-training before answer selection.}
Relation ranking improves on Direct for all eight of our models and all
six released baselines. The gain ranges from $0.17$ points for our Looped
\attnres{} decoder to $17.97$ for GPT-2. This pattern is consistent with
transfer from a task that already uses the downstream five-way scoring rule.
Description modeling alone improves all four of our encoders but lowers
accuracy for three of our four decoders. For the Looped \attnres{} encoder,
adding descriptions before relation ranking raises accuracy from $49.23\%$
to $50.36\%$, its best result. For the corresponding decoder, the same
change takes accuracy from $46.58\%$ to $46.25\%$. These comparisons use
the Dev-selected budget for each route (\cref{app:selection}).

\paragraph{Comparison at a smaller pretraining scale.}
Our models make one pass over 3B FineWeb-Edu content tokens.
GPT-Neo-125M saw 300B tokens over the course of pretraining on The Pile
\citep{eleuther2021gptneo}. With roughly one hundredth of that reported
token exposure, both of our looped decoders score higher than GPT-Neo
on every matched route. The Looped Standard decoder, for example, reaches
$47.70\%$ after $\stageL\!\rightarrow\!\stageR$, compared with $45.12\%$
for GPT-Neo. GPT-2 illustrates how the comparison changes with
post-training. Our Looped Standard decoder leads on Direct ($45.45\%$
versus $31.83\%$), then trails after relation ranking ($47.46\%$ versus
$49.80\%$). OPT-125M leads our decoders on all four routes.
The released encoders likewise retain their lead. On its strongest
route, our Looped \attnres{} encoder scores $50.36\%$, compared with
$53.18\%$ for BERT-base on that route. These comparisons establish which
checkpoints our models surpass on IHTest. Differences in corpus composition,
tokenization, and training compute limit what they imply about data efficiency.

\subsection{Pretraining Loss and Compute Cost}
\label{sec:ablations}

Depth reads
lower held-out loss for both backbones, at both traversal counts. For the
looped encoder, masked-token cross-entropy falls from $1.7090$ to $1.6484$.
For the looped decoder, next-token cross-entropy falls from $3.2886$ to
$3.2589$. Thus a decoder can assign higher likelihood to held-out text
while seeing little improvement in its ability to choose among answers.
\Cref{app:losses} gives the losses for each training objective.

\paragraph{What does access to depth history cost?}
At four traversals, depth reads increase the parameter count by
$0.03\%$. Even after summarization, each read scores and combines a growing
list of token representations, while an additive residual updates one
running state. Over the full 3B-token pretraining schedule, depth reads increase
training and evaluation cost in equivalent H100 hours by
$87\%$ for the encoder and $94\%$ for the decoder relative to their
additive controls (\cref{tab:runtime}). Peak allocated memory increases
by a factor of $3.5$ in both backbones (\cref{tab:resource-profile}).
At this additional cost, Direct accuracy improves by about ten percentage
points for the encoder and one point for the decoder.

\subsection{Discussion}
\label{sec:limitations}

The final EOS token sees the full question and candidate in both backbones.
A question token inside the encoder can already incorporate information
from the candidate. The decoder processes the same question token before
it sees the candidate. One possibility is that the encoder gains more from
rereading question-token representations because they already reflect the
particular answer being scored.
The comparison also pairs bidirectional attention with MLM and causal
attention with CLM. Varying visibility and pretraining supervision separately,
or intervening on the depth sources available to individual tokens, would
help distinguish these explanations.

We report accuracy on one CommonsenseQA split using the seed and
training protocol in \cref{app:config}. The Direct comparison holds the
fine-tuning budget fixed. The staged routes introduce additional training
and offer more budgets for development-set selection. The definition
corpora differ in whether they cover held-out concepts, and the version
is unresolved for some configurations (\cref{app:data}). The staged results
therefore reflect differences in exposure and search as well as supervision.
A separate architectural question is how much detail earlier loops need
to retain. Comparing sums with learned summaries or full cross-loop history
at the same traversal count would test that choice directly.

\section{Conclusion}
\label{sec:conclusion}

In our decision model, an individual sublayer output remains directly
accessible until its loop ends. Its contribution then travels forward in a
tokenwise sum that later layers can read separately from the summaries of
other loops. Evaluating this design through one answer-scoring rule reveals a difference between the two backbones.
Repeated layers improve answer selection in both. The additional gain from
reading depth history is more consistent for the encoder. Across both families, training on
relation-derived choices is the most consistent way to transfer the
additional knowledge supervision to the decision task.

\subsection*{AI use statement}
Generative AI assisted with initial drafting, source inspection, and
language editing.

\subsection*{Reproducibility statement}
\Cref{sec:method} specifies the forward pass and training objectives.
\Cref{app:config,app:objectives,app:data} give the model dimensions,
optimization settings, and dataset splits. \Cref{app:selection,app:accuracy}
report the epoch budgets and paired Dev and IHTest accuracies for each model.
\label{main:lastpage}
\begingroup
\small
\bibliographystyle{arxiv_preprint}
\bibliography{references_looped}
\endgroup

\clearpage
\appendix
\setlength{\LTpre}{0.5\baselineskip}
\setlength{\LTpost}{0.5\baselineskip}
\captionsetup[table]{skip=6pt}

\section{Architecture and Training Configuration}
\label{app:config}

\paragraph{Backbone configuration.}
Both backbones have 12 layers of width 768. Each layer has eight query
heads, four key/value heads, head dimension 96, and feed-forward width
2,432. We use dense feed-forward layers, rotary position embeddings with
base $10^6$, RMS normalization with $\epsilon=10^{-6}$, and zero dropout.
The RoBERTa vocabulary has 50,265 tokens and the context length is 512.
BOS, PAD, EOS, and MASK have indices 0, 1, 2, and 50,264, respectively.

{
\small
\begin{longtable}{@{}lrrrr@{}}
\caption{Optimization settings for the architecture comparison. Batch size
counts sequences for language modeling and questions for ranking, with
five sequences per question.}
\label{tab:training}\\
\toprule
Setting & Base PT & $\stageL$ & $\stageR$ & SFT \\
\midrule
\endfirsthead
\toprule
Setting & Base PT & $\stageL$ & $\stageR$ & SFT \\
\midrule
\endhead
\midrule
\endfoot
\bottomrule
\endlastfoot

Epoch budget & 1 & 1/2/3 & 1/2/3 & 5 \\
Peak learning rate & $3\!\times\!10^{-4}$ & $10^{-5}$ & $2\!\times\!10^{-5}$ & $2\!\times\!10^{-5}$ \\
Effective batch & 512 & 32 & 64 & 64 \\
Warmup fraction & 0.02 & 0.05 & 0.05 & 0.05 \\
AdamW $\beta_1$ & 0.9 & 0.9 & 0.9 & 0.9 \\
AdamW $\beta_2$ & 0.95 & 0.999 & 0.999 & 0.999 \\
Weight decay & 0.1 & 0.01 & 0.01 & 0.01 \\

\end{longtable}
}

AdamW uses $\epsilon=10^{-8}$ and a gradient clipping threshold of 1.0.
Learning rates increase linearly during warmup and then decay on a cosine
schedule to $10\%$ of the peak. The training protocol specifies BF16
precision and seed 42. Gradient accumulation maintains the effective batch
sizes in \cref{tab:training}, with smaller micro-batches for looped models.
Validation occurs every 500 updates during pretraining, after each epoch
of description modeling, and twice per epoch during relation ranking and SFT.

{
\small
\begin{longtable}{@{}lrrr@{}}
\caption{Language-model parameters by residual design and traversal count.
Counts are identical for the encoder and decoder. They include tied input
and output embeddings. The downstream scorer adds 768 parameters.}
\label{tab:architecture}\\
\toprule
Variant & Layers & Traversals & LM parameters \\
\midrule
\endfirsthead
\toprule
Variant & Layers & Traversals & LM parameters \\
\midrule
\endhead
\midrule
\endfoot
\bottomrule
\endlastfoot

Standard & 12 & 1 & 127,098,624 \\
Full \attnres{} & 12 & 1 & 127,137,024 \\
Looped Standard & 12 & 4 & 127,098,624 \\
Looped \attnres{} (\method) & 12 & 4 & 127,137,024 \\

\end{longtable}
}

For $D=12$, $T=4$, and $d=768$, the depth reads add 38,400 parameters.
Both looped models execute 48 layers per candidate. The depth-read model
combines 1,372 source vectors per token across those layers and the final
read, with at most 28 sources in any one read. Keeping all outputs
individually would raise those counts to 4,753 and 97, respectively
(\cref{eq:read-cost}).

\paragraph{Input processing.}
Both models mask padding keys. Each candidate is represented by its text
and the question, with answer letters and correctness labels excluded.
Training permutes the candidate order, and evaluation maps the selected
candidate to the original option order. Pairs exceeding the context limit
are excluded. The decoder scores each complete pair in one forward pass.

\subsection{Training Time and GPU Memory}
\label{app:resources}

We report the training and evaluation workload of each completed experiment
in equivalent hours on one NVIDIA H100 PCIe (80\,GB).
Measurements use BF16 and AdamW with four sequences per forward pass
for language modeling or four five-candidate questions for ranking.
Gradient accumulation preserves the effective batch sizes in
\cref{tab:training}.

\begin{center}
\begin{minipage}{\linewidth}
\captionsetup{type=table}
\caption{Training and evaluation cost.
Pretraining covers 3B content tokens. Brackets span the costs at the
fastest and slowest measured rates. Each
post-training column includes the Dev-selected adaptation budget
(\cref{tab:selected}) and five epochs of SFT after pretraining.}
\label{tab:runtime}
\centering\small
\setlength{\tabcolsep}{4pt}
\begin{tabular}{@{}lrr@{\hspace{9pt}}rrrr@{}}
\toprule
& \multicolumn{2}{c}{Pretraining} & \multicolumn{4}{c}{Selected post-training route} \\
\cmidrule(lr){2-3}\cmidrule(l){4-7}
Model & H100 h & Range & Direct & $\stageL$ & $\stageR$ & $\stageL\!\to\!\stageR$ \\
\midrule
Encoder Standard & 24.4 & [24.4, 24.7] & 0.24 & 0.26 & 0.64 & 0.40 \\
Encoder Full & 45.1 & [45.1, 45.2] & 0.34 & 0.39 & 0.91 & 0.77 \\
Encoder Looped Standard & 90.8 & [89.6, 91.4] & 0.66 & 0.74 & 1.77 & 1.43 \\
Encoder Looped AttnRes & 169.8 & [167.0, 174.0] & 1.08 & 1.13 & 1.67 & 2.95 \\
\midrule
Decoder Standard & 28.7 & [28.6, 28.8] & 0.23 & 0.26 & 0.37 & 0.51 \\
Decoder Full & 47.9 & [47.8, 48.3] & 0.33 & 0.35 & 0.53 & 0.94 \\
Decoder Looped Standard & 89.7 & [89.5, 90.3] & 0.63 & 0.66 & 1.69 & 1.37 \\
Decoder Looped AttnRes & 174.1 & [172.6, 178.4] & 0.94 & 1.00 & 1.57 & 2.36 \\
\bottomrule
\end{tabular}
\end{minipage}
\end{center}

\begin{center}
\begin{minipage}{\linewidth}
\captionsetup{type=table}
\caption{Measured pretraining throughput and memory. Throughput divides
the 262,144 input positions in a full optimizer update by its median
duration. Memory is peak live CUDA tensor allocation during the timed
updates, including optimizer state.}
\label{tab:resource-profile}
\centering\small
\begin{tabular}{@{}lrrr@{}}
\toprule
Model & Seconds/update & Input tokens/s & Peak allocated (GiB) \\
\midrule
Encoder Standard & 7.55 & 34,729 & 4.77 \\
Encoder Full & 13.94 & 18,810 & 9.64 \\
Encoder Looped Standard & 28.02 & 9,356 & 8.26 \\
Encoder Looped AttnRes & 52.48 & 4,995 & 28.78 \\
\midrule
Decoder Standard & 8.85 & 29,610 & 4.59 \\
Decoder Full & 14.78 & 17,737 & 9.46 \\
Decoder Looped Standard & 27.73 & 9,453 & 8.16 \\
Decoder Looped AttnRes & 53.85 & 4,868 & 28.69 \\
\bottomrule
\end{tabular}
\end{minipage}
\end{center}

\section{Knowledge Supervision and Objective Details}
\label{app:objectives}

\paragraph{Base language-model pretraining.}
Decoder pretraining minimizes next-token cross-entropy on valid targets.
Encoder pretraining minimizes cross-entropy at a sampled set $\mathcal M$ of
masked positions according to
\begin{align}
 \mathcal L_{\mathrm{CLM}}
   &= -\frac{1}{|\mathcal T_x|}\sum_{t\in\mathcal T_x}
      \log p_\theta(x_t\mid x_{<t}), \label{eq:clm}\\
 \mathcal L_{\mathrm{MLM}}
   &= -\frac{1}{|\mathcal M|}\sum_{t\in\mathcal M}
      \log p_\theta(x_t\mid\widetilde x). \label{eq:mlm}
\end{align}
Here $\mathcal T_x$ contains the valid next-token targets and $\widetilde x$
is the corrupted encoder input. We select $15\%$ of eligible content tokens,
excluding special tokens. Of the selected positions, $80\%$ receive MASK,
$10\%$ receive a random token, and $10\%$ keep their original token.
Training corruption varies by epoch. Evaluation fixes corruption by
sequence content and seed. We average each objective over its supervised
positions.

\paragraph{Negative sampling and knowledge partitions.}
For $(a,r,b)$, negatives are sampled from the observed tail pool for relation
$r$, excluding $a$ and all graph-known positive tails for $(a,r)$, including
positives absent from the smaller training corpus. Records with fewer than
four eligible negatives are skipped. A subject-based deterministic hash assigns
all facts with head $a$ to the same knowledge split. Since graph coverage
is incomplete, some alternative tails may also be valid answers.

\paragraph{Transfer and training branches.}
LM stages use a vocabulary head tied to input embeddings. Direct and
$\stageL$ routes initialize a new scalar scorer for SFT. Both $\stageR$ and
$\stageL\rightarrow\stageR$ transfer their trained scorer and backbone.
The independent stage budgets yield one Direct, three $\stageL$, three
$\stageR$, and nine combined configurations per model. Longer-budget runs
restart from their parent checkpoint and use their own full learning-rate
schedule. Staging therefore changes both supervision and the amount of
additional training.

\section{Datasets and Knowledge Coverage}
\label{app:data}

\paragraph{Question answering and relation ranking.}
The CommonsenseQA in-house split contains 8,500 training questions, 1,221
development questions, and 1,241 test questions. The relation corpus draws
on concepts and options from the training questions. It contains 25,695
training, 1,408 validation, and 1,859 test examples. We partition relation examples by head entity with target proportions of $90/5/5$. Differences in the number of
facts per entity account for the realized split sizes. Relation-ranking
validation and test examples are drawn from this corpus. Downstream
evaluation uses CommonsenseQA questions.

\paragraph{Definition and example corpora.}
The smaller corpus draws on training-set concepts and contains 2,134
records (1,920 train, 107 validation, 107 test). The expanded corpus
contains 16,588 records (14,929 train, 829 validation, 830 test). It
covers concepts and candidate terms from all question splits, including
held-out questions, and uses no answer labels. Definitions and examples
in the expanded corpus are deduplicated separately by normalized text.
The smaller corpus is documented for the looped training series, although
the version is known only for some individual configurations. Results involving $\stageL$ may therefore include exposure
to held-out terms during adaptation.

\section{Released Baselines and Pretraining Exposure}
\label{app:baselines}

{
\small\setlength{\tabcolsep}{5pt}
\begin{longtable}{@{}lrrrr@{}}
\caption{Released pretrained models. Parameter counts include the
language-model head. Context is the model's context limit and precision
is the numerical format used for adaptation and evaluation.}
\label{tab:baseline-config}\\
\toprule
Model & LM params (M) & Vocab rows & Context & Precision \\
\midrule
\endfirsthead
\toprule
Model & LM params (M) & Vocab rows & Context & Precision \\
\midrule
\endhead
\midrule
\endfoot
\bottomrule
\endlastfoot

BERT-base-uncased & 109.51 & 30,522 & 512 & BF16 \\
RoBERTa-base & 124.70 & 50,265 & 512 & BF16 \\
DeBERTa-base v1 & 139.24 & 50,265 & 512 & FP32 \\
GPT-2 & 124.44 & 50,257 & 1,024 & BF16 \\
GPT-Neo-125M & 125.20 & 50,257 & 2,048 & BF16 \\
OPT-125M & 125.24 & 50,272 & 2,048 & BF16 \\

\end{longtable}
}

\begin{table}[!htbp]
\centering
\caption{Pretraining exposure of released baselines. Corpus sizes retain
the units used by the original authors. Counts from the published schedules refer to input positions. B denotes $10^9$.}
\label{tab:baseline-exposure}
\small
\begin{tabularx}{\linewidth}{@{}l>{\raggedright\arraybackslash}X>{\raggedright\arraybackslash}X@{}}
\toprule
Model & Source corpus & Exposure / public budget \\
\midrule
BERT-base & Books and Wikipedia (3.3B words)
 & $\approx42.6$B positions (published schedule) \\
RoBERTa-base & Books, Wikipedia, news, web, stories (160GB)
 & $\approx2{,}100$B positions (published schedule) \\
DeBERTa-base v1 & Deduplicated books, Wikipedia, web, stories (78GB)
 & $\approx1{,}049$B positions (assuming length 512) \\
GPT-2 & WebText (40GB, over 8M documents)
 & Cumulative training tokens undisclosed \\
GPT-Neo-125M & The Pile (825\,GiB, \citealp{gao2020pile})
 & 300B tokens (model-card report) \\
OPT-125M & Deduplicated mixed corpus ($\approx180$B tokens)
 & 300B-token learning-rate schedule \\
\bottomrule
\end{tabularx}
\end{table}

\paragraph{Corpus size in tokens.}
For \cref{tab:main} we express the source text in billions of tokens before
any repetition during training. For BERT, RoBERTa, DeBERTa v1, and GPT-2,
we estimate GPT-2 BPE token equivalents from the reported text sizes of
16, 160, 78, and 40 decimal GB, respectively
\citep{liu2019roberta,he2021deberta,radford2019language}.
We use 0.24 tokens per UTF-8 byte as a prose approximation. For comparison,
\citet[Table~7]{gao2020pile} measure 0.2373 on English Wikipedia, 0.2430 on
BookCorpus2, and 0.2434 on OpenWebText2. The estimates depend on corpus
composition and can differ from counts under the original tokenizer,
especially BERT's WordPiece. For GPT-Neo we account for the Pile's mix
of prose, code, and other sources by multiplying each component's raw byte
size by its own GPT-2 token rate and summing across components
\citep[Tables~1 and~7]{gao2020pile}. This gives approximately 257B tokens
before upsampling, rounded to 260B in the table. OPT's roughly 180B tokens
come directly from the corpus report \citep{zhang2022opt}. For our models
we use the 3B content tokens counted by the RoBERTa tokenizer.

\paragraph{Token exposure under published schedules.}
We use the schedules and corpus sizes reported for the released models
\citep{devlin2019bert,liu2019roberta,he2021deberta,radford2019language,eleuther2021gptneo,zhang2022opt}.
BERT's length schedule gives
$256(900{,}000\cdot128+100{,}000\cdot512)\approx42.6$B positions.
The RoBERTa-base model card specifies 500K steps, an 8K-sequence batch, and
length 512, giving approximately 2.1T positions for 8,192 sequences per batch.
The DeBERTa count uses 1M steps, 2,048 sequences, and an assumed full
length of 512. Dynamic batching makes the realized token count variable.
GPT-Neo's model card reports 300B training tokens over 572,300 steps.
For OPT, the 300B figure describes the learning-rate schedule. The
controlled models in our study each see one epoch over 3B content tokens.

\section{Development Selection and Paired Results}
\label{app:selection}

For each model and route, we select the epoch budget with the highest
CSQA Dev accuracy at two-decimal precision. Ties favor fewer total
adaptation epochs, followed by lexicographic order of the $(L,R)$ budgets.
Within a run, ranking loss breaks ties in validation accuracy.

{
\small
\setlength{\tabcolsep}{4pt}
\begin{longtable}{@{}lrrrr@{}}
\caption{Dev and IHTest accuracy (\%) for the selected configuration
of each model and adaptation route on CommonsenseQA.}
\label{tab:selected-controlled}\\
\toprule
Model & Direct & $\stageL$ & $\stageR$ & $\stageL\!\rightarrow\!\stageR$ \\
\midrule
\endfirsthead
\toprule
Model & Direct & $\stageL$ & $\stageR$ & $\stageL\!\rightarrow\!\stageR$ \\
\midrule
\endhead
\midrule
\endfoot
\bottomrule
\endlastfoot

Decoder Standard & 45.54 / 41.50 & 45.45 / 40.69 & 47.58 / 43.84 & 47.91 / 41.02 \\
Encoder Standard & 33.82 / 32.15 & 42.26 / 37.07 & 41.11 / 38.36 & 45.54 / 38.84 \\
Decoder Full & 45.29 / 41.42 & 45.29 / 40.45 & 46.60 / 44.48 & 47.75 / 43.35 \\
Encoder Full & 45.86 / 41.02 & 45.62 / 42.55 & 48.40 / 44.96 & 49.88 / 44.96 \\
Decoder Looped Standard & 48.08 / 45.45 & 48.65 / 45.61 & 49.63 / 47.46 & 49.47 / 47.70 \\
Encoder Looped Standard & 37.59 / 35.46 & 48.16 / 43.92 & 45.45 / 39.56 & 48.08 / 44.00 \\
Decoder Looped AttnRes & 48.73 / 46.41 & 49.22 / 45.21 & 51.11 / 46.58 & 52.01 / 46.25 \\
Encoder Looped AttnRes & 51.19 / 45.69 & 51.60 / 46.66 & 52.25 / 49.23 & 53.15 / 50.36 \\

\end{longtable}
}

{
\small
\setlength{\tabcolsep}{5pt}
\begin{longtable}{@{}lrrrr@{}}
\caption{Released pretrained baselines. Cells show Dev / IHTest accuracy (\%)
at the selected configuration within each route in \cref{tab:main}.}
\label{tab:baselines}\\
\toprule
Model & Direct & $\stageL$ & $\stageR$ & $\stageL\!\rightarrow\!\stageR$ \\
\midrule
\endfirsthead
\toprule
Model & Direct & $\stageL$ & $\stageR$ & $\stageL\!\rightarrow\!\stageR$ \\
\midrule
\endhead
\midrule
\endfoot
\bottomrule
\endlastfoot

BERT-base & 52.09 / 50.12 & 52.33 / 50.68 & 57.90 / 52.78 & 58.15 / 53.18 \\
RoBERTa-base & 65.52 / 60.68 & 65.19 / 60.84 & 67.57 / 63.26 & 69.37 / 63.58 \\
DeBERTa-base & 66.09 / 61.89 & 66.91 / 62.29 & 69.37 / 63.90 & 69.70 / 62.93 \\
GPT-2 & 30.38 / 31.83 & 43.82 / 43.35 & 49.47 / 49.80 & 49.96 / 50.52 \\
GPT-Neo-125M & 44.96 / 43.51 & 45.29 / 43.35 & 48.32 / 43.76 & 47.83 / 45.12 \\
OPT-125M & 53.97 / 49.32 & 54.46 / 49.40 & 57.00 / 52.38 & 57.33 / 51.89 \\

\end{longtable}
}

{
\small
\begin{longtable}{@{}llll@{}}
\caption{Epoch budgets selected on CSQA Dev. Subscripts give the number
of epochs per knowledge stage before the common five-epoch SFT stage.}
\label{tab:selected}\\
\toprule
Model & $\stageL$ family & $\stageR$ family & $\stageL\!\rightarrow\!\stageR$ family \\
\midrule
\endfirsthead
\toprule
Model & $\stageL$ family & $\stageR$ family & $\stageL\!\rightarrow\!\stageR$ family \\
\midrule
\endhead
\midrule
\endfoot
\bottomrule
\endlastfoot

BERT-base & $\stageL_{3}$ & $\stageR_{2}$ & $\stageL_{3}\!\rightarrow\!\stageR_{3}$ \\
RoBERTa-base & $\stageL_{3}$ & $\stageR_{2}$ & $\stageL_{2}\!\rightarrow\!\stageR_{3}$ \\
DeBERTa-base & $\stageL_{2}$ & $\stageR_{3}$ & $\stageL_{3}\!\rightarrow\!\stageR_{2}$ \\
GPT-2 & $\stageL_{3}$ & $\stageR_{3}$ & $\stageL_{2}\!\rightarrow\!\stageR_{3}$ \\
GPT-Neo-125M & $\stageL_{3}$ & $\stageR_{3}$ & $\stageL_{1}\!\rightarrow\!\stageR_{3}$ \\
OPT-125M & $\stageL_{2}$ & $\stageR_{2}$ & $\stageL_{3}\!\rightarrow\!\stageR_{3}$ \\
Decoder Standard & $\stageL_{3}$ & $\stageR_{1}$ & $\stageL_{1}\!\rightarrow\!\stageR_{2}$ \\
Encoder Standard & $\stageL_{2}$ & $\stageR_{3}$ & $\stageL_{3}\!\rightarrow\!\stageR_{1}$ \\
Decoder Full & $\stageL_{1}$ & $\stageR_{1}$ & $\stageL_{2}\!\rightarrow\!\stageR_{3}$ \\
Encoder Full & $\stageL_{3}$ & $\stageR_{3}$ & $\stageL_{3}\!\rightarrow\!\stageR_{2}$ \\
Decoder Looped Standard & $\stageL_{1}$ & $\stageR_{3}$ & $\stageL_{1}\!\rightarrow\!\stageR_{2}$ \\
Encoder Looped Standard & $\stageL_{3}$ & $\stageR_{3}$ & $\stageL_{1}\!\rightarrow\!\stageR_{2}$ \\
Decoder Looped AttnRes & $\stageL_{1}$ & $\stageR_{1}$ & $\stageL_{3}\!\rightarrow\!\stageR_{2}$ \\
Encoder Looped AttnRes & $\stageL_{1}$ & $\stageR_{1}$ & $\stageL_{2}\!\rightarrow\!\stageR_{3}$ \\

\end{longtable}
}

For both Standard and Looped Standard decoders, the combined routes
$\stageL_1\rightarrow\stageR_2$ and $\stageL_2\rightarrow\stageR_2$ tie on
Dev. The shorter route is selected in both cases. For Looped \attnres{},
the selected combined route is $\stageL_2\rightarrow\stageR_3$ for the
encoder ($53.15\%$ Dev, $50.36\%$ IHTest) and
$\stageL_3\rightarrow\stageR_2$ for the decoder ($52.01\%$ Dev,
$46.25\%$ IHTest).

\section{Complete Downstream Accuracy Results}
\label{app:accuracy}

Each row corresponds to one training configuration. Each cell pairs the
Dev and IHTest accuracies of its Dev-selected SFT checkpoint.
{
\small\setlength{\tabcolsep}{4pt}
\begin{longtable}{@{}lrrr@{}}
\caption{Released encoder baselines on CommonsenseQA. Dev / IHTest accuracy (\%).}
\label{tab:baseline-all-enc}\\
\toprule
Route & BERT-base & RoBERTa-base & DeBERTa-base \\
\midrule
\endfirsthead
\toprule
Route & BERT-base & RoBERTa-base & DeBERTa-base \\
\midrule
\endhead
\midrule
\endfoot
\bottomrule
\endlastfoot

Direct & 52.09 / 50.12 & 65.52 / 60.68 & 66.09 / 61.89 \\
$\stageL_{1}$ & 51.19 / 51.09 & 65.11 / 62.93 & 66.58 / 62.37 \\
$\stageL_{2}$ & 51.43 / 49.72 & 64.86 / 61.48 & 66.91 / 62.29 \\
$\stageL_{3}$ & 52.33 / 50.68 & 65.19 / 60.84 & 66.42 / 62.45 \\
$\stageR_{1}$ & 55.69 / 52.30 & 67.40 / 63.42 & 67.65 / 63.58 \\
$\stageR_{2}$ & 57.90 / 52.78 & 67.57 / 63.26 & 68.30 / 63.74 \\
$\stageR_{3}$ & 57.33 / 52.70 & 67.49 / 63.66 & 69.37 / 63.90 \\
$\stageL_{1}\!\rightarrow\!\stageR_{1}$ & 57.00 / 51.57 & 66.91 / 63.09 & 67.57 / 63.74 \\
$\stageL_{1}\!\rightarrow\!\stageR_{2}$ & 57.74 / 53.10 & 68.06 / 63.98 & 68.55 / 63.50 \\
$\stageL_{1}\!\rightarrow\!\stageR_{3}$ & 57.25 / 53.99 & 69.21 / 63.66 & 69.12 / 63.90 \\
$\stageL_{2}\!\rightarrow\!\stageR_{1}$ & 56.02 / 51.49 & 66.99 / 62.61 & 67.98 / 63.09 \\
$\stageL_{2}\!\rightarrow\!\stageR_{2}$ & 57.90 / 52.30 & 69.21 / 63.42 & 69.12 / 63.82 \\
$\stageL_{2}\!\rightarrow\!\stageR_{3}$ & 58.07 / 52.94 & 69.37 / 63.58 & 69.45 / 64.71 \\
$\stageL_{3}\!\rightarrow\!\stageR_{1}$ & 55.86 / 51.89 & 67.16 / 63.74 & 68.14 / 62.53 \\
$\stageL_{3}\!\rightarrow\!\stageR_{2}$ & 57.82 / 52.94 & 68.71 / 64.38 & 69.70 / 62.93 \\
$\stageL_{3}\!\rightarrow\!\stageR_{3}$ & 58.15 / 53.18 & 68.80 / 63.01 & 69.29 / 65.11 \\

\end{longtable}
}

{
\small\setlength{\tabcolsep}{4pt}
\begin{longtable}{@{}lrrr@{}}
\caption{Released decoder baselines on CommonsenseQA. Dev / IHTest accuracy (\%).}
\label{tab:baseline-all-dec}\\
\toprule
Route & GPT-2 & GPT-Neo-125M & OPT-125M \\
\midrule
\endfirsthead
\toprule
Route & GPT-2 & GPT-Neo-125M & OPT-125M \\
\midrule
\endhead
\midrule
\endfoot
\bottomrule
\endlastfoot

Direct & 30.38 / 31.83 & 44.96 / 43.51 & 53.97 / 49.32 \\
$\stageL_{1}$ & 41.85 / 42.87 & 44.88 / 43.19 & 53.97 / 49.48 \\
$\stageL_{2}$ & 43.41 / 43.76 & 45.05 / 42.47 & 54.46 / 49.40 \\
$\stageL_{3}$ & 43.82 / 43.35 & 45.29 / 43.35 & 53.56 / 50.20 \\
$\stageR_{1}$ & 49.06 / 47.95 & 47.09 / 43.67 & 56.10 / 51.25 \\
$\stageR_{2}$ & 49.39 / 49.72 & 48.08 / 45.53 & 57.00 / 52.38 \\
$\stageR_{3}$ & 49.47 / 49.80 & 48.32 / 43.76 & 56.59 / 51.97 \\
$\stageL_{1}\!\rightarrow\!\stageR_{1}$ & 47.42 / 48.03 & 46.11 / 43.92 & 56.76 / 52.05 \\
$\stageL_{1}\!\rightarrow\!\stageR_{2}$ & 49.30 / 48.67 & 46.76 / 45.93 & 57.17 / 53.51 \\
$\stageL_{1}\!\rightarrow\!\stageR_{3}$ & 49.30 / 50.36 & 47.83 / 45.12 & 56.18 / 52.70 \\
$\stageL_{2}\!\rightarrow\!\stageR_{1}$ & 47.83 / 47.62 & 46.27 / 43.43 & 57.25 / 51.89 \\
$\stageL_{2}\!\rightarrow\!\stageR_{2}$ & 49.14 / 49.07 & 46.36 / 44.40 & 56.59 / 52.38 \\
$\stageL_{2}\!\rightarrow\!\stageR_{3}$ & 49.96 / 50.52 & 46.60 / 44.64 & 57.17 / 53.67 \\
$\stageL_{3}\!\rightarrow\!\stageR_{1}$ & 47.50 / 47.86 & 45.70 / 44.72 & 57.08 / 50.52 \\
$\stageL_{3}\!\rightarrow\!\stageR_{2}$ & 49.30 / 49.72 & 46.68 / 44.72 & 56.18 / 52.38 \\
$\stageL_{3}\!\rightarrow\!\stageR_{3}$ & 48.98 / 50.12 & 46.93 / 44.72 & 57.33 / 51.89 \\

\end{longtable}
}

{
\small\setlength{\tabcolsep}{4pt}
\begin{longtable}{@{}lrrrr@{}}
\caption{Single-pass common-backbone models on CommonsenseQA. Dev / IHTest accuracy (\%). Dec. and Enc. denote decoder and encoder.}
\label{tab:controlled-all}\\
\toprule
Route & Dec. Standard & Enc. Standard & Dec. Full & Enc. Full \\
\midrule
\endfirsthead
\toprule
Route & Dec. Standard & Enc. Standard & Dec. Full & Enc. Full \\
\midrule
\endhead
\midrule
\endfoot
\bottomrule
\endlastfoot

Direct & 45.54 / 41.50 & 33.82 / 32.15 & 45.29 / 41.42 & 45.86 / 41.02 \\
$\stageL_{1}$ & 45.29 / 40.85 & 40.79 / 36.99 & 45.29 / 40.45 & 45.13 / 40.29 \\
$\stageL_{2}$ & 44.80 / 40.53 & 42.26 / 37.07 & 43.73 / 41.42 & 44.80 / 41.58 \\
$\stageL_{3}$ & 45.45 / 40.69 & 41.85 / 38.28 & 44.47 / 41.50 & 45.62 / 42.55 \\
$\stageR_{1}$ & 47.58 / 43.84 & 39.89 / 36.58 & 46.60 / 44.48 & 48.32 / 42.79 \\
$\stageR_{2}$ & 46.93 / 42.79 & 40.87 / 37.63 & 45.13 / 41.50 & 48.32 / 43.67 \\
$\stageR_{3}$ & 46.03 / 39.08 & 41.11 / 38.36 & 46.03 / 42.87 & 48.40 / 44.96 \\
$\stageL_{1}\!\rightarrow\!\stageR_{1}$ & 46.93 / 43.59 & 43.90 / 40.85 & 46.03 / 42.22 & 49.22 / 45.04 \\
$\stageL_{1}\!\rightarrow\!\stageR_{2}$ & 47.91 / 41.02 & 43.24 / 39.81 & 46.11 / 43.84 & 49.39 / 46.17 \\
$\stageL_{1}\!\rightarrow\!\stageR_{3}$ & 46.68 / 40.05 & 43.90 / 39.40 & 45.95 / 44.00 & 48.65 / 44.56 \\
$\stageL_{2}\!\rightarrow\!\stageR_{1}$ & 46.93 / 42.95 & 43.73 / 41.42 & 46.44 / 44.48 & 49.55 / 45.21 \\
$\stageL_{2}\!\rightarrow\!\stageR_{2}$ & 47.91 / 41.34 & 43.65 / 40.85 & 46.76 / 42.47 & 49.06 / 45.45 \\
$\stageL_{2}\!\rightarrow\!\stageR_{3}$ & 46.36 / 40.69 & 45.13 / 41.98 & 47.75 / 43.35 & 49.14 / 45.37 \\
$\stageL_{3}\!\rightarrow\!\stageR_{1}$ & 47.42 / 41.82 & 45.54 / 38.84 & 45.62 / 45.37 & 49.55 / 44.72 \\
$\stageL_{3}\!\rightarrow\!\stageR_{2}$ & 46.76 / 41.50 & 44.96 / 39.89 & 46.44 / 43.67 & 49.88 / 44.96 \\
$\stageL_{3}\!\rightarrow\!\stageR_{3}$ & 47.01 / 41.50 & 45.13 / 40.37 & 46.93 / 43.92 & 48.16 / 46.01 \\

\end{longtable}
}

\label{app:looped-results}
{
\small
\setlength{\tabcolsep}{4pt}
\begin{longtable}{@{}lrrrr@{}}
\caption{Looped models on CommonsenseQA. Each cell gives Dev / IHTest
accuracy (\%) for one adaptation route and epoch budget.}
\label{tab:looped-all}\\
\toprule
 & \multicolumn{2}{c}{Looped Standard} & \multicolumn{2}{c}{Looped \attnres{}} \\
\cmidrule(lr){2-3}\cmidrule(l){4-5}
Route & Decoder & Encoder & Decoder & Encoder \\
\midrule
\endfirsthead
\toprule
 & \multicolumn{2}{c}{Looped Standard} & \multicolumn{2}{c}{Looped \attnres{}} \\
\cmidrule(lr){2-3}\cmidrule(l){4-5}
Route & Decoder & Encoder & Decoder & Encoder \\
\midrule
\endhead
\midrule
\endfoot
\bottomrule
\endlastfoot

Direct & 48.08 / 45.45 & 37.59 / 35.46 & 48.73 / 46.41 & 51.19 / 45.69 \\
$\stageL_{1}$ & 48.65 / 45.61 & 45.95 / 40.93 & 49.22 / 45.21 & 51.60 / 46.66 \\
$\stageL_{2}$ & 48.32 / 47.22 & 47.09 / 43.27 & 48.24 / 48.03 & 51.02 / 48.35 \\
$\stageL_{3}$ & 48.32 / 44.32 & 48.16 / 43.92 & 48.65 / 46.66 & 51.35 / 46.25 \\
$\stageR_{1}$ & 48.32 / 47.38 & 43.00 / 39.40 & 51.11 / 46.58 & 52.25 / 49.23 \\
$\stageR_{2}$ & 49.30 / 44.96 & 44.14 / 40.69 & 50.70 / 43.92 & 51.19 / 49.15 \\
$\stageR_{3}$ & 49.63 / 47.46 & 45.45 / 39.56 & 49.88 / 44.08 & 51.52 / 48.51 \\
$\stageL_{1}\!\rightarrow\!\stageR_{1}$ & 48.73 / 44.72 & 47.17 / 43.51 & 51.43 / 47.95 & 52.33 / 49.15 \\
$\stageL_{1}\!\rightarrow\!\stageR_{2}$ & 49.47 / 47.70 & 48.08 / 44.00 & 50.61 / 49.72 & 52.33 / 49.23 \\
$\stageL_{1}\!\rightarrow\!\stageR_{3}$ & 48.89 / 45.45 & 46.85 / 45.77 & 49.96 / 48.59 & 50.70 / 49.15 \\
$\stageL_{2}\!\rightarrow\!\stageR_{1}$ & 49.22 / 44.72 & 47.09 / 45.45 & 51.76 / 47.46 & 52.33 / 49.48 \\
$\stageL_{2}\!\rightarrow\!\stageR_{2}$ & 49.47 / 47.54 & 47.26 / 44.56 & 50.61 / 46.25 & 52.83 / 50.68 \\
$\stageL_{2}\!\rightarrow\!\stageR_{3}$ & 48.73 / 45.21 & 46.85 / 46.98 & 49.30 / 46.25 & 53.15 / 50.36 \\
$\stageL_{3}\!\rightarrow\!\stageR_{1}$ & 48.98 / 43.84 & 47.26 / 43.51 & 51.60 / 48.43 & 52.01 / 49.64 \\
$\stageL_{3}\!\rightarrow\!\stageR_{2}$ & 48.89 / 48.43 & 47.67 / 44.32 & 52.01 / 46.25 & 52.42 / 47.86 \\
$\stageL_{3}\!\rightarrow\!\stageR_{3}$ & 48.48 / 45.85 & 47.01 / 47.95 & 51.11 / 47.14 & 52.01 / 48.51 \\

\end{longtable}
}

\section{Architecture Contrasts}
\label{app:contrasts}

We subtract the IHTest accuracies of the two named variants within each
backbone and adaptation route. Each variant uses its own Dev-selected
budget. Direct compares models trained with the same fine-tuning budget.
{
\small
\setlength{\tabcolsep}{3pt}
\begin{longtable}{@{}llrrrr@{}}
\caption{Differences in IHTest accuracy, in percentage points. Std denotes
additive residuals and Full denotes single-pass Attention Residuals.}
\label{tab:deltas}\\
\toprule
Comparison & Architecture & Direct & $\stageL$ & $\stageR$ & $\stageL\!\rightarrow\!\stageR$ \\
\midrule
\endfirsthead
\toprule
Comparison & Architecture & Direct & $\stageL$ & $\stageR$ & $\stageL\!\rightarrow\!\stageR$ \\
\midrule
\endhead
\midrule
\endfoot
\bottomrule
\endlastfoot

Full $-$ Standard & Decoder & $-0.08$ & $-0.24$ & $+0.64$ & $+2.33$ \\
Full $-$ Standard & Encoder & $+8.87$ & $+5.48$ & $+6.60$ & $+6.12$ \\
Looped Std $-$ Standard & Decoder & $+3.95$ & $+4.92$ & $+3.62$ & $+6.68$ \\
Looped Std $-$ Standard & Encoder & $+3.31$ & $+6.85$ & $+1.20$ & $+5.16$ \\
Looped AttnRes $-$ Looped Std & Decoder & $+0.96$ & $-0.40$ & $-0.88$ & $-1.45$ \\
Looped AttnRes $-$ Looped Std & Encoder & $+10.23$ & $+2.74$ & $+9.67$ & $+6.36$ \\
Looped AttnRes $-$ Full & Decoder & $+4.99$ & $+4.76$ & $+2.10$ & $+2.90$ \\
Looped AttnRes $-$ Full & Encoder & $+4.67$ & $+4.11$ & $+4.27$ & $+5.40$ \\

\end{longtable}
}

\paragraph{How the depth-read advantage changes with looping.}
Let $A$ denote IHTest accuracy for a given backbone and route. We compute
\begin{equation}
 I=(A_{\mathrm{loop,attn}}-A_{\mathrm{loop,std}})
   -(A_{\mathrm{single,attn}}-A_{\mathrm{single,std}}).
 \label{eq:interaction}
\end{equation}
For direct fine-tuning, the depth-read advantage grows by
$10.23-8.87=1.36$ points in the encoder and $0.96-(-0.08)=1.04$ points in the
decoder. The looped depth-read variant adds a summary for each completed
pass. This contrast describes the change from the single-pass to the
four-pass model under that summary rule.

\section{Pretraining, Knowledge Adaptation, and SFT Losses}
\label{app:losses}

We report language-model losses per supervised token and ranking losses
per question. Each comparison uses the same objective and data split.

\subsection{Evaluation Metrics}
\label{app:perplexity}

\paragraph{Causal language modeling.}
For $N$ scored next-token targets, the token-averaged negative log-likelihood
in nats is
\begin{equation}
 H_{\mathrm{CLM}}=-\frac{1}{N}\sum_{i=1}^{N}
       \ln p_\theta(x_i\mid x_{<i}),\qquad
 \mathrm{PPL}=\exp(H_{\mathrm{CLM}}),\qquad
 \ln\mathrm{PPL}=H_{\mathrm{CLM}}.
 \label{eq:logppl}
\end{equation}
We weight each batch by its number of supervised tokens. The looped decoder
evaluations each score 6,519,148 held-out next-token targets.

\paragraph{Masked language modeling.}
The looped encoders each score 976,076 selected token positions under the
corruption protocol in \cref{app:objectives}. We report their masked-token
cross-entropy. For comparison, leave-one-out pseudo-perplexity
\citep{salazar2020masked} uses a separate masked sequence for each token and is defined by
\begin{equation}
 \ln\mathrm{PPPL}(x)=-\frac{1}{n}\sum_{i=1}^{n}
       \ln p_\theta(x_i\mid x_{\setminus i}).
 \label{eq:pppl}
\end{equation}
Here $x_{\setminus i}$ masks only position $i$. The exponential of our
sampled-mask loss uses the corruption distribution of MLM.

\paragraph{Five-candidate ranking.}
The ranking loss is cross-entropy per question,
\[
 H_{\mathrm{rank}}=-N_q^{-1}\sum_i\ln p(y_i\mid q_i,\mathcal C_i).
\]
This loss measures the probability assigned to the labeled answer within
the five candidates. Accuracy measures how often that answer receives
the highest score.

{
\small
\begin{longtable}{@{}llrr@{}}
\caption{Held-out pretraining cross-entropy and its exponential for the
looped models. For the decoder the exponential is causal perplexity. For
the encoder it is exponentiated masked-token cross-entropy.}
\label{tab:logged-loss}\\
\toprule
Model & Objective & Test CE & $\exp(\mathrm{CE})$ \\
\midrule
\endfirsthead
\toprule
Model & Objective & Test CE & $\exp(\mathrm{CE})$ \\
\midrule
\endhead
\bottomrule
\endlastfoot
Decoder Looped Standard & CLM & 3.288576 & 26.804657 \\
Encoder Looped Standard & MLM & 1.709034 & 5.523624 \\
Decoder Looped AttnRes & CLM & 3.258875 & 26.020239 \\
Encoder Looped AttnRes & MLM & 1.648424 & 5.198778 \\
\end{longtable}
}

\subsection{Base Pretraining and Definition/Example Adaptation}

{
\small
\begin{longtable}{@{}llrr@{}}
\caption{Language-model loss after one epoch on 3B pretraining tokens.
Validation and test values come from the checkpoint with minimum validation
loss. The decoder uses next-token loss and the encoder uses masked-token loss.}
\label{tab:pretraining-all}\\
\toprule
Model & Objective & Validation & Test \\
\midrule
\endfirsthead
\toprule
Model & Objective & Validation & Test \\
\midrule
\endhead
\midrule
\endfoot
\bottomrule
\endlastfoot

Decoder Standard & CLM & 3.3140 & 3.3275 \\
Decoder Full & CLM & 3.2699 & 3.2837 \\
Encoder Standard & MLM & 1.7704 & 1.7893 \\
Encoder Full & MLM & 1.7272 & 1.7463 \\
\midrule
Decoder Looped Standard & CLM & 3.2749 & 3.2886 \\
Decoder Looped AttnRes & CLM & 3.2461 & 3.2589 \\
Encoder Looped Standard & MLM & 1.6891 & 1.7090 \\
Encoder Looped AttnRes & MLM & 1.6301 & 1.6484 \\

\end{longtable}
}

Longer description-modeling budgets reduce validation and test loss for
all four single-pass variants. The corresponding answer-selection accuracies
in \cref{tab:controlled-all} vary across budgets.
{
\small\setlength{\tabcolsep}{4pt}
\begin{longtable}{@{}lrrrr@{}}
\caption{Definition/example LM adaptation for single-pass models. Each cell gives validation / test loss at the checkpoint with minimum validation loss.}
\label{tab:controlled-lm}\\
\toprule
Route & Dec. Standard & Enc. Standard & Dec. Full & Enc. Full \\
\midrule
\endfirsthead
\toprule
Route & Dec. Standard & Enc. Standard & Dec. Full & Enc. Full \\
\midrule
\endhead
\midrule
\endfoot
\bottomrule
\endlastfoot

$\stageL_{1}$ & 2.2976 / 2.3721 & 1.5370 / 1.7073 & 2.2638 / 2.3486 & 1.3285 / 1.5235 \\
$\stageL_{2}$ & 2.2315 / 2.3033 & 1.4493 / 1.5984 & 2.2060 / 2.2843 & 1.2579 / 1.4426 \\
$\stageL_{3}$ & 2.2054 / 2.2765 & 1.3849 / 1.5336 & 2.1821 / 2.2629 & 1.2122 / 1.4007 \\

\end{longtable}
}

{
\small\setlength{\tabcolsep}{4pt}
\begin{longtable}{@{}lrrr@{}}
\caption{Released encoder baselines after definition/example adaptation. Each cell gives validation / test loss at the checkpoint with minimum validation loss. Each model uses its own tokenizer.}
\label{tab:baseline-lm-enc}\\
\toprule
Route & BERT-base & RoBERTa-base & DeBERTa-base \\
\midrule
\endfirsthead
\toprule
Route & BERT-base & RoBERTa-base & DeBERTa-base \\
\midrule
\endhead
\midrule
\endfoot
\bottomrule
\endlastfoot

$\stageL_{1}$ & 1.4157 / 1.4328 & 0.9714 / 0.8785 & 0.7856 / 1.1867 \\
$\stageL_{2}$ & 1.2947 / 1.3119 & 0.9045 / 0.8361 & 0.7395 / 1.1112 \\
$\stageL_{3}$ & 1.2155 / 1.2495 & 0.8657 / 0.8208 & 0.7062 / 1.0774 \\

\end{longtable}
}

{
\small\setlength{\tabcolsep}{4pt}
\begin{longtable}{@{}lrrr@{}}
\caption{Released decoder baselines after definition/example adaptation. Each cell gives validation / test loss at the checkpoint with minimum validation loss. Each model uses its own tokenizer.}
\label{tab:baseline-lm-dec}\\
\toprule
Route & GPT-2 & GPT-Neo-125M & OPT-125M \\
\midrule
\endfirsthead
\toprule
Route & GPT-2 & GPT-Neo-125M & OPT-125M \\
\midrule
\endhead
\midrule
\endfoot
\bottomrule
\endlastfoot

$\stageL_{1}$ & 2.4518 / 2.4943 & 2.7105 / 2.8306 & 2.2801 / 2.3433 \\
$\stageL_{2}$ & 2.3460 / 2.3873 & 2.3964 / 2.4942 & 2.1963 / 2.2554 \\
$\stageL_{3}$ & 2.2922 / 2.3376 & 2.3288 / 2.4180 & 2.1547 / 2.2122 \\

\end{longtable}
}

\subsection{Knowledge-Relation Ranking}

For released checkpoints, each cell gives the minimum validation loss over
training and the knowledge-test loss at the checkpoint with highest
knowledge-validation accuracy. These two quantities may come from different
checkpoints. For the single-pass controlled models, we report the
validation loss at the accuracy-selected checkpoint.
{
\small\setlength{\tabcolsep}{4pt}
\begin{longtable}{@{}lrrr@{}}
\caption{Relation ranking for released encoders. Minimum validation loss / test loss at the accuracy-selected checkpoint.}
\label{tab:baseline-rank-enc}\\
\toprule
Route & BERT-base & RoBERTa-base & DeBERTa-base \\
\midrule
\endfirsthead
\toprule
Route & BERT-base & RoBERTa-base & DeBERTa-base \\
\midrule
\endhead
\midrule
\endfoot
\bottomrule
\endlastfoot

$\stageR_{1}$ & 0.5687 / 0.9636 & 0.4734 / 0.8633 & 0.4710 / 0.8624 \\
$\stageR_{2}$ & 0.5367 / 1.0442 & 0.4417 / 0.9534 & 0.4417 / 0.9567 \\
$\stageR_{3}$ & 0.5201 / 1.2102 & 0.4278 / 0.9406 & 0.4432 / 1.0589 \\
$\stageL_{1}\!\rightarrow\!\stageR_{1}$ & 0.5606 / 0.9653 & 0.4802 / 0.8914 & 0.4727 / 0.8772 \\
$\stageL_{1}\!\rightarrow\!\stageR_{2}$ & 0.5293 / 1.0441 & 0.4397 / 0.9551 & 0.4370 / 0.9645 \\
$\stageL_{1}\!\rightarrow\!\stageR_{3}$ & 0.5177 / 1.0445 & 0.4271 / 0.9954 & 0.4398 / 1.0244 \\
$\stageL_{2}\!\rightarrow\!\stageR_{1}$ & 0.5595 / 0.9626 & 0.4790 / 0.8901 & 0.4702 / 0.8668 \\
$\stageL_{2}\!\rightarrow\!\stageR_{2}$ & 0.5279 / 1.0348 & 0.4417 / 0.9588 & 0.4344 / 0.9685 \\
$\stageL_{2}\!\rightarrow\!\stageR_{3}$ & 0.5118 / 1.0280 & 0.4267 / 1.0408 & 0.4314 / 0.9905 \\
$\stageL_{3}\!\rightarrow\!\stageR_{1}$ & 0.5587 / 0.9559 & 0.4798 / 0.8939 & 0.4674 / 0.8839 \\
$\stageL_{3}\!\rightarrow\!\stageR_{2}$ & 0.5302 / 1.0368 & 0.4404 / 0.9477 & 0.4256 / 0.9398 \\
$\stageL_{3}\!\rightarrow\!\stageR_{3}$ & 0.5083 / 1.1985 & 0.4298 / 1.0528 & 0.4370 / 1.0027 \\

\end{longtable}
}

{
\small\setlength{\tabcolsep}{4pt}
\begin{longtable}{@{}lrrr@{}}
\caption{Relation ranking for released decoders. Minimum validation loss / test loss at the accuracy-selected checkpoint.}
\label{tab:baseline-rank-dec}\\
\toprule
Route & GPT-2 & GPT-Neo-125M & OPT-125M \\
\midrule
\endfirsthead
\toprule
Route & GPT-2 & GPT-Neo-125M & OPT-125M \\
\midrule
\endhead
\midrule
\endfoot
\bottomrule
\endlastfoot

$\stageR_{1}$ & 0.7787 / 0.9718 & 0.7283 / 0.9730 & 0.5308 / 0.8661 \\
$\stageR_{2}$ & 0.6883 / 0.9309 & 0.6711 / 1.0027 & 0.5266 / 1.0688 \\
$\stageR_{3}$ & 0.6534 / 0.9195 & 0.6612 / 1.1342 & 0.5338 / 1.0180 \\
$\stageL_{1}\!\rightarrow\!\stageR_{1}$ & 0.7558 / 0.9593 & 0.7377 / 0.9833 & 0.5350 / 0.8576 \\
$\stageL_{1}\!\rightarrow\!\stageR_{2}$ & 0.6891 / 0.9401 & 0.6748 / 1.0165 & 0.5260 / 1.0413 \\
$\stageL_{1}\!\rightarrow\!\stageR_{3}$ & 0.6543 / 0.9262 & 0.6704 / 1.1875 & 0.5431 / 1.0159 \\
$\stageL_{2}\!\rightarrow\!\stageR_{1}$ & 0.7609 / 0.9552 & 0.7275 / 0.9817 & 0.5385 / 0.8540 \\
$\stageL_{2}\!\rightarrow\!\stageR_{2}$ & 0.6919 / 0.9429 & 0.6723 / 1.0206 & 0.5276 / 1.0280 \\
$\stageL_{2}\!\rightarrow\!\stageR_{3}$ & 0.6546 / 0.9184 & 0.6674 / 1.2067 & 0.5329 / 1.3958 \\
$\stageL_{3}\!\rightarrow\!\stageR_{1}$ & 0.7597 / 0.9503 & 0.7214 / 0.9833 & 0.5340 / 0.8610 \\
$\stageL_{3}\!\rightarrow\!\stageR_{2}$ & 0.6892 / 0.9376 & 0.6679 / 1.0161 & 0.5310 / 1.0285 \\
$\stageL_{3}\!\rightarrow\!\stageR_{3}$ & 0.6570 / 0.9355 & 0.6664 / 1.1980 & 0.5351 / 1.0740 \\

\end{longtable}
}

{
\small\setlength{\tabcolsep}{4pt}
\begin{longtable}{@{}lrrrr@{}}
\caption{Validation loss for single-pass relation ranking at the accuracy-selected checkpoint.}
\label{tab:controlled-rank}\\
\toprule
Route & Dec. Standard & Enc. Standard & Dec. Full & Enc. Full \\
\midrule
\endfirsthead
\toprule
Route & Dec. Standard & Enc. Standard & Dec. Full & Enc. Full \\
\midrule
\endhead
\midrule
\endfoot
\bottomrule
\endlastfoot

$\stageR_{1}$ & 0.6981 & 0.7875 & 0.6774 & 0.7028 \\
$\stageR_{2}$ & 0.7859 & 0.7791 & 0.7937 & 0.7203 \\
$\stageR_{3}$ & 1.5972 & 0.7764 & 0.7792 & 0.7098 \\
$\stageL_{1}\!\rightarrow\!\stageR_{1}$ & 0.6870 & 0.7381 & 0.6729 & 0.6806 \\
$\stageL_{1}\!\rightarrow\!\stageR_{2}$ & 0.7984 & 0.7383 & 0.8123 & 0.7287 \\
$\stageL_{1}\!\rightarrow\!\stageR_{3}$ & 0.8038 & 0.7615 & 0.7266 & 0.7086 \\
$\stageL_{2}\!\rightarrow\!\stageR_{1}$ & 0.6872 & 0.7447 & 0.6695 & 0.6790 \\
$\stageL_{2}\!\rightarrow\!\stageR_{2}$ & 0.7936 & 0.7223 & 0.7852 & 0.7357 \\
$\stageL_{2}\!\rightarrow\!\stageR_{3}$ & 0.8057 & 0.7530 & 0.7249 & 0.7125 \\
$\stageL_{3}\!\rightarrow\!\stageR_{1}$ & 0.6848 & 0.7473 & 0.6685 & 0.6835 \\
$\stageL_{3}\!\rightarrow\!\stageR_{2}$ & 0.7908 & 0.7288 & 0.8041 & 0.6872 \\
$\stageL_{3}\!\rightarrow\!\stageR_{3}$ & 0.8032 & 0.7506 & 0.7285 & 0.7126 \\

\end{longtable}
}

For instance, the Standard decoder trained for three relation-ranking epochs
reaches a minimum validation loss of $0.7395$. The checkpoint selected by
validation accuracy has loss $1.5972$.

\subsection{Downstream Ranking}

We evaluate downstream ranking loss at the checkpoint selected by
CommonsenseQA Dev accuracy. The tables cover the single-pass models and
the additive looped models.
{
\small\setlength{\tabcolsep}{4pt}
\begin{longtable}{@{}lrrrr@{}}
\caption{Single-pass SFT. Dev / IHTest ranking loss at the same Dev-accuracy-selected checkpoint.}
\label{tab:controlled-sft-loss}\\
\toprule
Route & Dec. Standard & Enc. Standard & Dec. Full & Enc. Full \\
\midrule
\endfirsthead
\toprule
Route & Dec. Standard & Enc. Standard & Dec. Full & Enc. Full \\
\midrule
\endhead
\midrule
\endfoot
\bottomrule
\endlastfoot

Direct & 1.4978 / 1.6066 & 1.5629 / 1.5528 & 1.4492 / 1.4875 & 1.3296 / 1.3951 \\
$\stageL_{1}$ & 2.3620 / 2.6208 & 1.9690 / 2.1772 & 4.5255 / 4.8194 & 1.8355 / 1.9696 \\
$\stageL_{2}$ & 2.3537 / 2.6052 & 3.0338 / 3.3350 & 1.4727 / 1.5147 & 1.8555 / 1.9771 \\
$\stageL_{3}$ & 2.4101 / 2.6535 & 2.9942 / 3.2486 & 2.4961 / 2.5137 & 1.3786 / 1.4313 \\
$\stageR_{1}$ & 1.4610 / 1.5363 & 1.5357 / 1.5842 & 1.5508 / 1.5445 & 1.3723 / 1.4889 \\
$\stageR_{2}$ & 1.2953 / 1.3268 & 1.5108 / 1.5780 & 1.5147 / 1.5635 & 1.2684 / 1.3155 \\
$\stageR_{3}$ & 3.5394 / 3.8909 & 2.4409 / 2.5492 & 1.4984 / 1.5701 & 1.4266 / 1.5230 \\
$\stageL_{1}\!\rightarrow\!\stageR_{1}$ & 1.4872 / 1.5769 & 1.5019 / 1.5094 & 3.8458 / 4.1058 & 1.4105 / 1.4625 \\
$\stageL_{1}\!\rightarrow\!\stageR_{2}$ & 1.4876 / 1.6304 & 2.5051 / 2.5882 & 1.2780 / 1.2923 & 1.4485 / 1.5226 \\
$\stageL_{1}\!\rightarrow\!\stageR_{3}$ & 1.5023 / 1.6469 & 3.8260 / 3.9699 & 1.5595 / 1.6143 & 1.4187 / 1.5174 \\
$\stageL_{2}\!\rightarrow\!\stageR_{1}$ & 1.4248 / 1.5052 & 1.4598 / 1.4796 & 1.2806 / 1.2850 & 1.4412 / 1.4930 \\
$\stageL_{2}\!\rightarrow\!\stageR_{2}$ & 1.4976 / 1.6284 & 1.5206 / 1.5712 & 1.5626 / 1.6247 & 1.4386 / 1.5219 \\
$\stageL_{2}\!\rightarrow\!\stageR_{3}$ & 1.5073 / 1.6454 & 1.5132 / 1.5561 & 1.5621 / 1.6448 & 1.4078 / 1.5351 \\
$\stageL_{3}\!\rightarrow\!\stageR_{1}$ & 1.5137 / 1.5843 & 3.5393 / 3.6323 & 1.2846 / 1.2876 & 1.4403 / 1.5060 \\
$\stageL_{3}\!\rightarrow\!\stageR_{2}$ & 1.5130 / 1.6417 & 1.5378 / 1.5633 & 1.5772 / 1.5988 & 1.3998 / 1.5080 \\
$\stageL_{3}\!\rightarrow\!\stageR_{3}$ & 1.5103 / 1.6498 & 1.5079 / 1.5786 & 1.2687 / 1.3027 & 1.4224 / 1.5242 \\

\end{longtable}
}

{
\small
\begin{longtable}{@{}lrr@{}}
\caption{Looped Standard ranking loss on CSQA Dev / IHTest at the
accuracy-selected checkpoints in \cref{tab:looped-all}.}
\label{tab:looped-loss}\\
\toprule
Route & Decoder Looped Standard & Encoder Looped Standard \\
\midrule
\endfirsthead
\toprule
Route & Decoder Looped Standard & Encoder Looped Standard \\
\midrule
\endhead
\midrule
\endfoot
\bottomrule
\endlastfoot

Direct & 1.347993 / 1.444467 & 1.862705 / 1.998183 \\
$\stageL_{1}$ & 2.077542 / 2.274822 & 1.760725 / 1.894215 \\
$\stageL_{2}$ & 1.418860 / 1.523006 & 1.308521 / 1.374834 \\
$\stageL_{3}$ & 2.138913 / 2.348020 & 1.323978 / 1.363240 \\
$\stageR_{1}$ & 1.245355 / 1.272025 & 1.471652 / 1.552473 \\
$\stageR_{2}$ & 1.397943 / 1.539610 & 1.495845 / 1.586926 \\
$\stageR_{3}$ & 1.227700 / 1.258735 & 3.264465 / 3.669245 \\
$\stageL_{1}\!\rightarrow\!\stageR_{1}$ & 1.540949 / 1.653624 & 1.366923 / 1.466611 \\
$\stageL_{1}\!\rightarrow\!\stageR_{2}$ & 1.229449 / 1.267712 & 1.409950 / 1.525588 \\
$\stageL_{1}\!\rightarrow\!\stageR_{3}$ & 2.105874 / 2.315618 & 1.461073 / 1.522049 \\
$\stageL_{2}\!\rightarrow\!\stageR_{1}$ & 2.354075 / 2.635754 & 1.435073 / 1.501897 \\
$\stageL_{2}\!\rightarrow\!\stageR_{2}$ & 1.231208 / 1.265531 & 1.480902 / 1.564993 \\
$\stageL_{2}\!\rightarrow\!\stageR_{3}$ & 1.408677 / 1.542362 & 1.245086 / 1.277841 \\
$\stageL_{3}\!\rightarrow\!\stageR_{1}$ & 1.396071 / 1.515358 & 3.185065 / 3.518560 \\
$\stageL_{3}\!\rightarrow\!\stageR_{2}$ & 1.237765 / 1.262531 & 1.434412 / 1.512842 \\
$\stageL_{3}\!\rightarrow\!\stageR_{3}$ & 2.116687 / 2.281620 & 1.240556 / 1.276021 \\

\end{longtable}
}

\end{document}